\documentclass[10pt,twocolumn,letterpaper]{article}
\usepackage{cvpr}              
\usepackage{tabularx}
\usepackage{multirow}
\usepackage{booktabs}
\usepackage{xcolor}
\usepackage[table,xcdraw]{xcolor}
\usepackage{array}

\usepackage{tabularx}
\usepackage{caption}
\usepackage{graphicx}

\definecolor{cvprblue}{rgb}{0.21,0.49,0.74}
\usepackage[pagebackref,breaklinks,colorlinks,allcolors=cvprblue]{hyperref}

\def\paperID{****} 
\def\confName{CVPR}
\def\confYear{2026}

\title{High-Quality and Efficient Turbulence Mitigation with Events}

\author{Xiaoran Zhang\textsuperscript{1}\footnotemark[2], Jian Ding\textsuperscript{1}\footnotemark[2], Yuxing Duan\textsuperscript{1}, Haoyue Liu\textsuperscript{1}\footnotemark[1], Gang Chen\textsuperscript{2}, Yi Chang\textsuperscript{1}, Luxin Yan\textsuperscript{1}\\
	\textsuperscript{1} State Key Laboratory of Multispectral Information Intelligent Processing Technology\\
	School of Artificial Intelligence and Automation, Huazhong University of Science and Technology\\
	\textsuperscript{2} School of Computer Science and Engineering, Sun Yat-sen University \\
	{\tt\small \{xiaoranzhang, dingjian, liuhy, yichang\}@hust.edu.cn, cheng83@mail.sysu.edu.cn}}
	
\begin{document}
\maketitle
{
	\renewcommand{\thefootnote}%
	{\fnsymbol{footnote}}
	\footnotetext[0]{$\dagger$Equal contribution. *Corresponding author.} 
}

\begin{abstract}
Turbulence mitigation (TM) is highly ill-posed due to the stochastic nature of atmospheric turbulence. Most methods rely on multiple frames recorded by conventional cameras to capture stable patterns in natural scenarios. However, they inevitably suffer from a trade-off between accuracy and efficiency: more frames enhance restoration at the cost of higher system latency and larger data overhead. Event cameras, equipped with microsecond temporal resolution and efficient sensing of dynamic changes, offer an opportunity to break the bottleneck. In this work, we present EHETM, a high-quality and efficient TM method inspired by the superiority of events to model motions in continuous sequences. We discover two key phenomena: (1) turbulence-induced events exhibit distinct polarity alternation correlated with sharp image gradients, providing structural cues for restoring scenes; and (2) dynamic objects form spatiotemporally coherent ``event tubes'' in contrast to irregular patterns within turbulent events, providing motion priors for disentangling objects from turbulence. Based on these insights, we design two complementary modules that respectively leverage polarity-weighted gradients for scene refinement and event-tube constraints for motion decoupling, achieving high-quality restoration with few frames. Furthermore, we construct two real-world event-frame turbulence datasets covering atmospheric and thermal cases. Experiments show that EHETM outperforms SOTA methods, especially under scenes with dynamic objects, while reducing data overhead and system latency by approximately 77.3\% and 89.5\%, respectively. Our code is available at \url{https://github.com/Xavier667/EHETM}.
\end{abstract}    
\section{Introduction}
\label{sec:Introduction}

Atmospheric and thermal turbulence are major sources of degradation in long-range imaging, arising from random fluctuations of air refractive index \cite{jin2021neutralizing, guo2024direct, hill2025deep, anantrasirichai2013atmospheric, chan2022tilt}. Such turbulence, introducing complex distortions characterized by geometric tilts and spatial-varying blurs, degrades image quality and hinders the performance of downstream vision tasks \cite{cornett2023expanding, zhang2024source}.

\begin{figure}[t]
	\setlength{\abovecaptionskip}{5pt}
	\setlength{\belowcaptionskip}{-3pt}
	\centering
	\includegraphics[width=\linewidth]{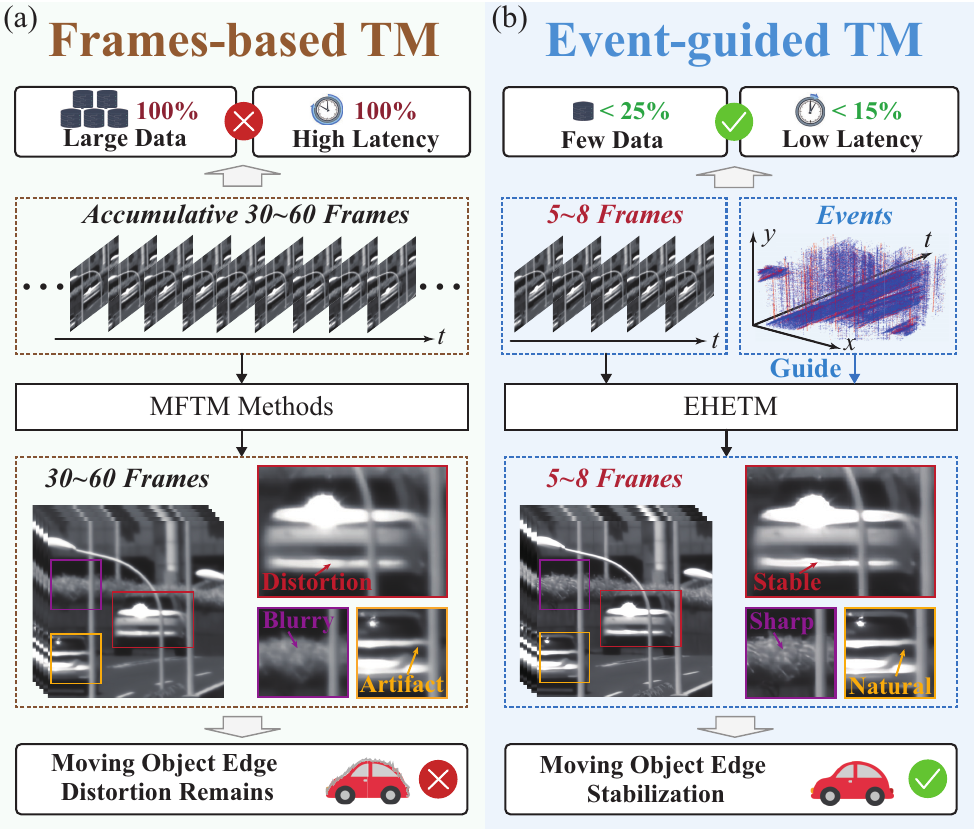}
	\caption{Workflow comparison between our EHETM and multi-frame TM methods (MFTM, exemplified by DATUM~\cite{zhang2024spatio}).
		(a) DATUM~\cite{zhang2024spatio} suffers from high system latency and large overhead due to multi-frame accumulation, while distortions at dynamic object edges remains.
		(b) Our EHETM uses events for guidance, producing higher-quality and efficient restoration with few frames.}
	\label{fig:supp-1}
\end{figure}

To address this challenge, early classical methods based on pixel registration, lucky fusion, and deblurring, etc., have provided insights into turbulence mitigation (TM) \cite{anantrasirichai2013atmospheric, hirsch2010efficient, patel2019adaptive, caliskan2014atmospheric, fazlali2022atmospheric, zhu2012removing, qin2024unsupervised, xu2024long}. However, they are computationally expensive and struggling with dynamic scenes. Recently, deep learning-based methods have shown great potential, which can be categorized into single-frame \cite{jaiswal2023physics, lau2021atfacegan, mao2022single, mei2023ltt, nair2021confidence, nair2023ddpm, rai2022removing, yasarla2022cnn, lau2021semi, chen2019u, yasarla2021learning, wu2024semi, saha2024turb} and multi-frame methods \cite{anantrasirichai2023atmospheric, wang2023revelation, gao2019atmospheric, chak2018subsampled, jin2021neutralizing, fazlali2022atmospheric, xia2025planet, feng2023turbugan, liu2023eaf, mao2020image, zhang2024imaging, zhang2024spatio, zou2024deturb, wang2024real, cai2024temporally, zhang2025learning, zhang2024physical, li2021unsupervised, jiang2023nert}. Single-frame methods offer simplicity but are limited by the ill-posedness of turbulence degradation, undermining restoration quality. Multi-frame methods utilize sufficient temporal cues from a large number of frames (\textit{typically 30--60}) recorded by conventional cameras. However, as illustrated in Fig.~\ref{fig:supp-1} (a), they suffer a trade-off: more frames enhance restoration at the expense of higher system latency\footnotemark (\textit{about 1--2~s}) and larger data overhead (\textit{about 30--60$\times$}), impeding their real-time applicability. \noindent \footnotetext{System latency: the output for a given frame may depend on future inputs, which imposes an inherent delay independent of computation.}
 
Beyond the efficiency–accuracy trade-off, conventional cameras suffer from low-frequency sampling, failing to offer adequate and stable motion information for accurate motion decoupling between object motion and turbulence. It raises a question: \textit{with a high-speed imaging system capable of capturing transient turbulence dynamics, can we extract finer-grained temporal cues within a shorter duration to break the bottleneck in frames-only methods and achieve high-quality and efficient restoration?}

\begin{figure}[t]
	\setlength{\abovecaptionskip}{5pt}
	\setlength{\belowcaptionskip}{-3pt}
	\centering
	\includegraphics[width=\linewidth]{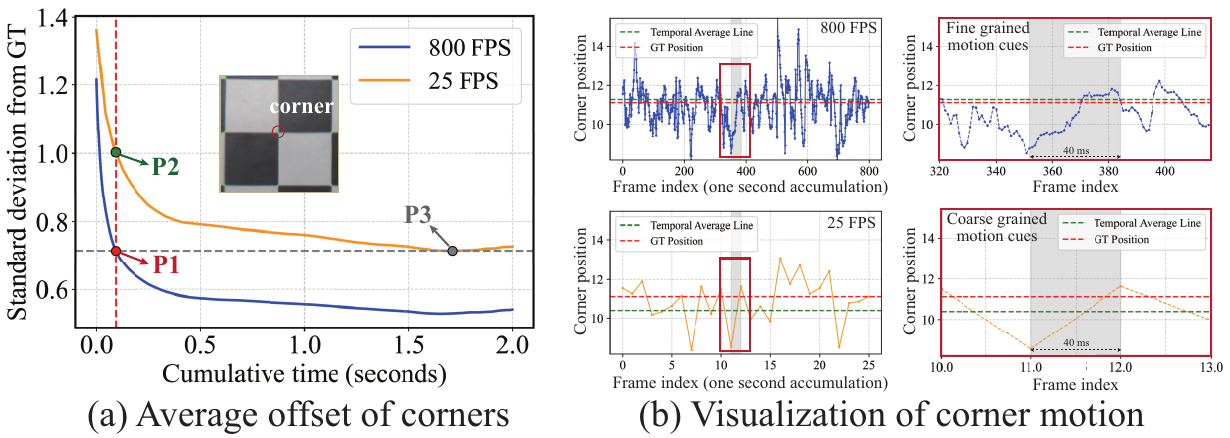}
	\caption{Merits of high-speed observations. (a) It requires less accumulation time to reach the same error (P1 vs. P3) and achieves lower error within the same time (P1 vs. P2). (b) It captures finer-gained motion cues compared to low-temporal-resolution ones.}
	\label{fig:supp-2}
\end{figure}

Event cameras \cite{gallego2020event, liu2024seeing, yu2024eventps} asynchronously sense per-pixel brightness changes, capturing motion cues with microsecond temporal resolution. Such properties make them a powerful complement to frames-based TM methods by offering temporally dense yet spatially efficient (edge-triggered) information. To intuitively illustrate advantages of high-temporal-resolution data, Fig.~\ref{fig:supp-2} compares temporal averaging from a high-speed (800 FPS) and a conventional (25 FPS) camera under thermal turbulence. By capturing fine-grained motion cues, high-temporal-resolution observations rapidly and accurately reveal the zero-mean characteristics of turbulence, producing higher quality within a shorter time than low-temporal-resolution ones. However, high-speed frame cameras incur prohibitively high recourse consumption, making long-duration operation impractical. 

In this work, we propose an Event-guided High-quality and Efficient TM (EHETM) method. As shown in Fig.~\ref{fig:supp-1} (b), EHETM addresses two key limitations of frames-only methods: it \textbf{accelerates TM} by supplying high-temporal-resolution information from events, and it \textbf{enhances TM quality} by exploiting temporally dense events to accurately decouple object motion from turbulence. We discover two phenomena: (1) Turbulence-induced pixel displacement exhibits polarity alternation of events along sharp image gradients, offering structural cues for restoring scenes. (2) Objects in natural scenes exhibits continuous motion, which form spatiotemporally coherent ``event tubes" in contrast to the irregular patterns of turbulent events, providing stable priors for decoupling object motion from turbulence. Motivated by these observations, we design two complementary modules in EHETM: the former encodes polarity alternation frequency into gradients to reduce edge distortions of scenes, while the latter stabilizes dynamic objects by modeling event tubes. They jointly learn stable patterns from turbulence for high-quality and efficient TM.

Besides, for comprehensive demonstration, we construct two real-world event-frame paired turbulence datasets: Close-range Thermal Turbulence Hybrid (CTTH) dataset and Long-range Atmospheric Turbulence Hybrid (LATH) dataset. CTTH contains dynamic objects with GTs, while LATH covers diverse scenes at varying shooting distances. In summary, we make the following primary contributions:

\begin{itemize}
	\setlength\itemsep{0.5em} 
	\item We break the bottleneck induced by accuracy–efficiency trade-off in frames-based approaches by exploiting fine-grained motion cues from events within a short duration. With only 5 to 8 input frames and events, our method significantly reduces data overhead and system latency, while achieving high-quality restoration.
	
	\item We reveal distinct spatiotemporal characteristics of turbulent events via experimental and theoretical analysis---scenes exhibit polarity alternation while dynamic objects form event tubes. We accordingly propose EHETM that leverage polarity-weighted gradients for scene refinement and event-tube priors for motion decoupling, attaining high-quality and efficient TM with few frames.
	
	\item We bridge the data gap in TM by constructing the first long-range atmospheric turbulence event-frame dataset shooting at various distances, and the first thermal turbulence event-frame dataset with dynamic objects and GTs. Extensive experiments show that our method outperforms SOTA methods in both restoration quality and efficiency.
\end{itemize}

\begin{figure*}[t]
	\setlength{\abovecaptionskip}{5pt}
	\setlength{\belowcaptionskip}{-3pt}
	\centering
	\includegraphics[width=\linewidth]{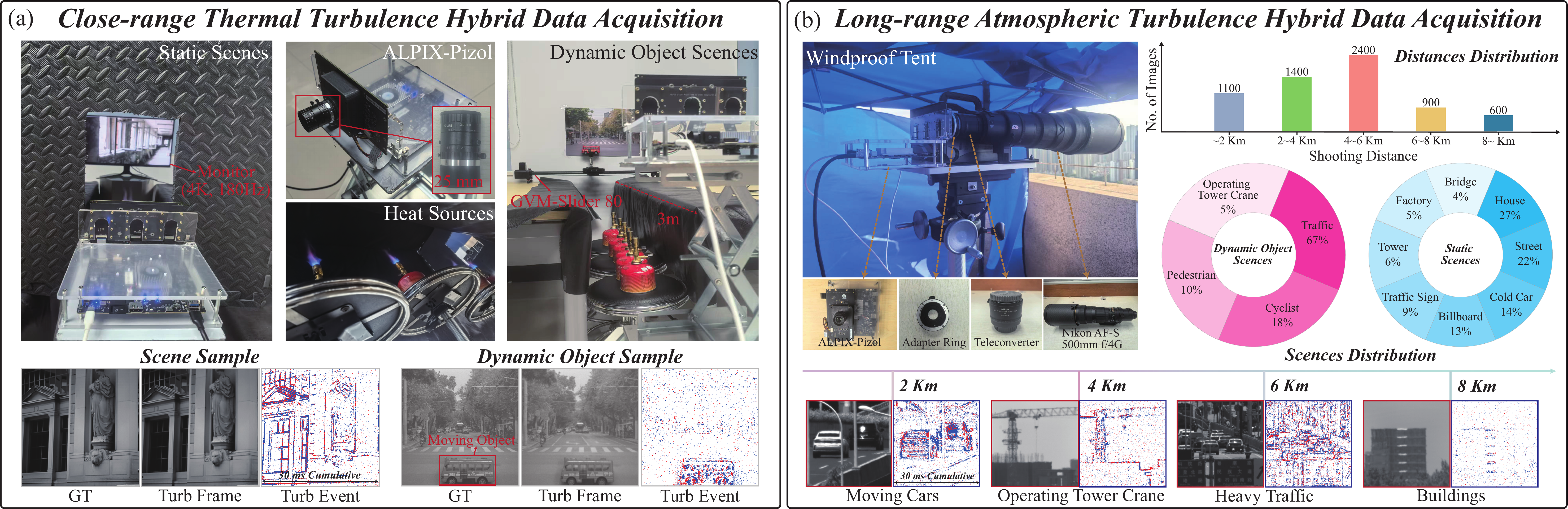}
	\caption{Our real-world event-frame turbulence datasets and acquisition setups. (a) CTTH dataset is captured under adjustable thermal turbulence, containing about 30k image-event pairs with dynamic objects and GTs. (b) LATH dataset is acquired using a 1000 mm lens covering multiple scenes at various shooting distances.}
	\label{fig:supp-6}
\end{figure*}

\section{Related Works}
\label{sec:RelatedWork}
\crefname{table}{Table}{Tables} 
\bibliographystyle{unsrt}

\textbf{Learning-based Turbulence Mitigation.} Recent AI advances have pushed learning-based methods to the forefront of TM, covering both single- and multi-frame paradigms. Single-frame methods, including CNNs \cite{chen2019u, yasarla2022cnn, yasarla2021learning}, GANs \cite{rai2022removing, lau2021atfacegan, mei2023ltt}, physics-informed Transformers \cite{mao2022single, jaiswal2023physics}, diffusion-based methods \cite{nair2023ddpm}, and approaches exploiting narrowband priors \cite{xia2024nb} or the seg-then-res pipeline \cite{saha2024turb}, etc., offer simplicity but lack temporal information, limiting restoration performance. Multi-frame methods exploit temporal cues, including video-based GANs \cite{jin2021neutralizing, feng2023turbugan} and Transformers \cite{zhang2024imaging}, network-incorporated traditional paradigms \cite{zhang2024spatio}, the state-space model \cite{zhang2025learning}, and neural representations \cite{li2021unsupervised, jiang2023nert, cai2024temporally}, etc., showing strong potential for TM. However, this benefit comes at a cost: every extra frame brings additional system latency and data overhead. Moreover, at low frame rates, severe aliasing between object motion and turbulence are difficult to be accurately decoupled, degrading restoration quality. In contrast, our method overcomes these limitations by leveraging high-speed events to guide few frames for efficient and high-quality restoration.

\noindent\textbf{Event-based Video Restoration.} Event cameras can efficiently capture transient motion often missed by conventional cameras \cite{liu2025timetracker, xie2025diffusion, xiao2025event, duan2024led, zhou2024bring}. In video super-resolution and motion deblurring, events provide precise motion cues, improving edge recovery \cite{xiao2024event, jing2021turning, kai2023video, pan2019bringing, yu2024learning, zhang2023generalizing, kim2024cmta}. These results highlight that high-temporal-resolution events reveal motion patterns invisible to conventional cameras, offering new opportunities for TM. Recent works have explored event-based TM: Liu \textit{et al}. \cite{liu2025evturb} model tilt and blur effects in turbulent events, while Li \textit{et al}. \cite{li2025egtm} are committed to identify “event-lucky regions”. Although insightful, these studies do not exploit events on objects, leaving motion-turbulence entanglement largely unaddressed. We address this gap by leveraging fine-grained temporal information to model object motion.

\noindent\textbf{Turbulence Datasets.} Current datasets primarily encompass three categories: physics-based simulation algorithms \cite{zhang2024spatio, zhang2024imaging, jin2021neutralizing, wang2023revelation}, close-range thermal turbulence \cite{mao2022single, ug22023, anantrasirichai2013atmospheric, jin2021neutralizing, liu2025evturb, li2025egtm}, and long-range atmospheric turbulence \cite{hirsch2010efficient, gilles2017open, jin2021neutralizing, wang2023revelation, mao2020image, xu2024long}. Most existing datasets consist of low-frame-rate image sequences, with only \cite{liu2025evturb, li2025egtm} providing event-frame paired data for thermal turbulence. To our best knowledge, there is currently no event-frame paired turbulence dataset that includes dynamic objects with GTs, limiting the validation of event-based methods under dynamic object scenes. Moreover, the lack of real-world hybrid atmospheric turbulence datasets hinders the evaluation of event-based methods across diverse turbulence conditions. Thus, we construct CTTH and LATH datasets to bridge the data gap.

\section{Real-World Hybrid Turbulence Datasets}
\label{sec:Dataset}
\crefname{table}{Table}{Tables} 
\bibliographystyle{unsrt}

\noindent Capturing real-world event–frame paired turbulence data is very valuable, as high-speed events can precisely characterize the spatiotemporal dynamics of turbulence, providing finer-gained motion cues than conventional frames.

\noindent\textbf{CTTH Dataset.} The CTTH setup is shown in Fig.~\ref{fig:supp-6} (a). Inspired by \cite{jin2021neutralizing}, we use a 4K 180Hz monitor and an ALPIX-pizol hybrid camera with a 25mm lens. The camera’s split-focal-plane design ensures hardware-level synchronization of events and images, eliminating spatiotemporal misalignment. Our base data source comprises over 300 groups of clean images (512 × 512) from TMT dataset \cite{zhang2024imaging}, spanning indoor, outdoor, textual, and facial categories, which are super-resolved to 2048 × 2048 via Real-ESRGAN \cite{wang2021realesrgan}. During acquisition, the synchronized monitor-camera setup captured 100 frames per scene at 25~fps, producing approximately 30000 turbulent and turbulence-free image-event pairs (512 × 512). Turbulence is generated via multiple gas sources, with the camera placed 3 meters from the monitor. For dynamic object scenes, the setup incorporates multiple posters, a programmable slider (GVM-Slider), mechanical fixtures, and the ALPIX-pizol camera. Various toy vehicles in controlled experiments across different poster backgrounds. Precise regulation of slider motion parameters ensures spatial consistency, with 200 frames captured per scene at 25 fps, producing approximately 3000 turbulent and turbulence-free image-event pairs (512 × 512).

\noindent\textbf{LATH Dataset.} As shown in Fig.~\ref{fig:supp-6} (b), LATH dataset is captured using the ALPIX-pizol hybrid camera configured with a 1000 mm telephoto setup (500 mm lens + 2× teleconverter) at 25 fps across multiple distances and scenes. To ensure stability, all recordings are conducted inside a windproof tent with a professional tripod to eliminate vibrations. Each sequence lasts about 8 seconds, from which the stable 5 seconds segment is retained, yielding roughly 6500 curated image-event pairs at 512 × 512 resolution. LATH dataset covers shooting distances from 1-8 km and diverse long-range scenes, such as moving cars, operating tower cranes, heavy traffic, and buildings, etc.
\section{Turbulent Event Signal Characteristics}
\label{sec:Characteristics}
\crefname{table}{Table}{Tables} 
\bibliographystyle{unsrt}

\begin{figure}[t]
	\setlength{\abovecaptionskip}{5pt}
	\setlength{\belowcaptionskip}{-3pt}
	\centering
	\includegraphics[width=\linewidth]{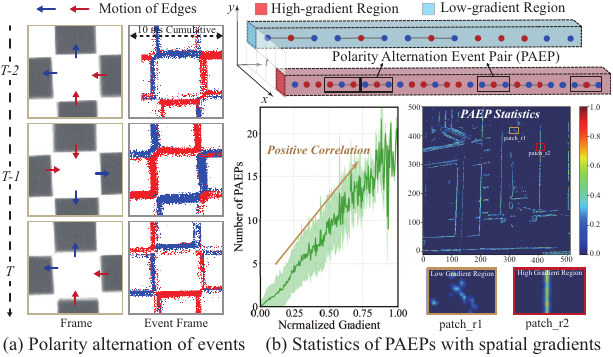}
	\caption{Event responses of scenes caused by turbulence. (a) Turbulence induces event polarity alternation along structural edges. (b) The PAEP numbers correlate with sharp image gradients, providing structural cues for suppressing edge distortions.}
	\label{fig:supp-3}
\end{figure}

\subsection{Turb-induced Scene Event Response} 
\label{sec:Polarity}
In image domain, turbulence mainly distorts gradient-rich regions, causing geometric tilts and edge blurring. In event domain, it remains concentrated in these areas, manifesting as high-frequency polarity alternation induced by local brightness oscillations, as presented in Fig.~\ref{fig:supp-3} (a).

To quantify this relationship, we conduct statistical experiments under varying thermal turbulence conditions. The analysis of polarity alternation event pair (PAEP) counts versus sharp image gradients reveals a strong positive correlation across different turbulence levels (mean Pearson's $r = 0.82$, $p < 0.001$), as presented in the line graph in Fig.~\ref{fig:supp-3} (b). This demonstrates that regions with higher gradients experience more intense turbulence-induced brightness fluctuations, resulting in more frequent polarity alternations. Notably, we discover that edges generated by high frequency PAEPs at high-gradient locations tend to be sharper, since these regions are more sensitive to abrupt intensity variations and produce denser responses along structural boundaries, leading to better preservation of fine details. We attribute this phenomenon to the zero-mean property of turbulence, which causes perturbations to predominantly oscillate around their original positions \cite{xu2024long}.

\begin{figure}[t]
	\setlength{\abovecaptionskip}{5pt}
	\setlength{\belowcaptionskip}{-5pt}
	\centering
	\includegraphics[width=\linewidth]{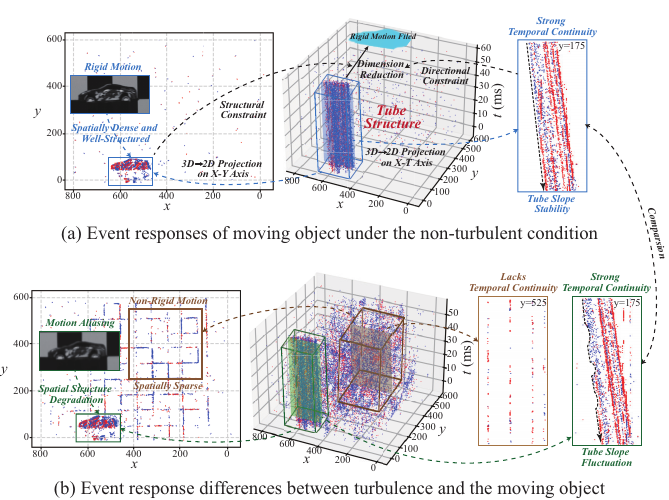}
	\caption{Event responses of the dynamic object under turbulence. (a) Turbulence-free: the object triggers temporally continuous and spatially dense events, forming a stable event tube. (b) Under turbulence: the event tube is structurally degraded, while turbulence produces temporally discontinuous and spatially sparse events.}
	\label{fig:supp-4}
\end{figure}

\begin{figure*}[t]
	\setlength{\abovecaptionskip}{5pt}
	\setlength{\belowcaptionskip}{-3pt}
	\centering
	\includegraphics[width=\linewidth]{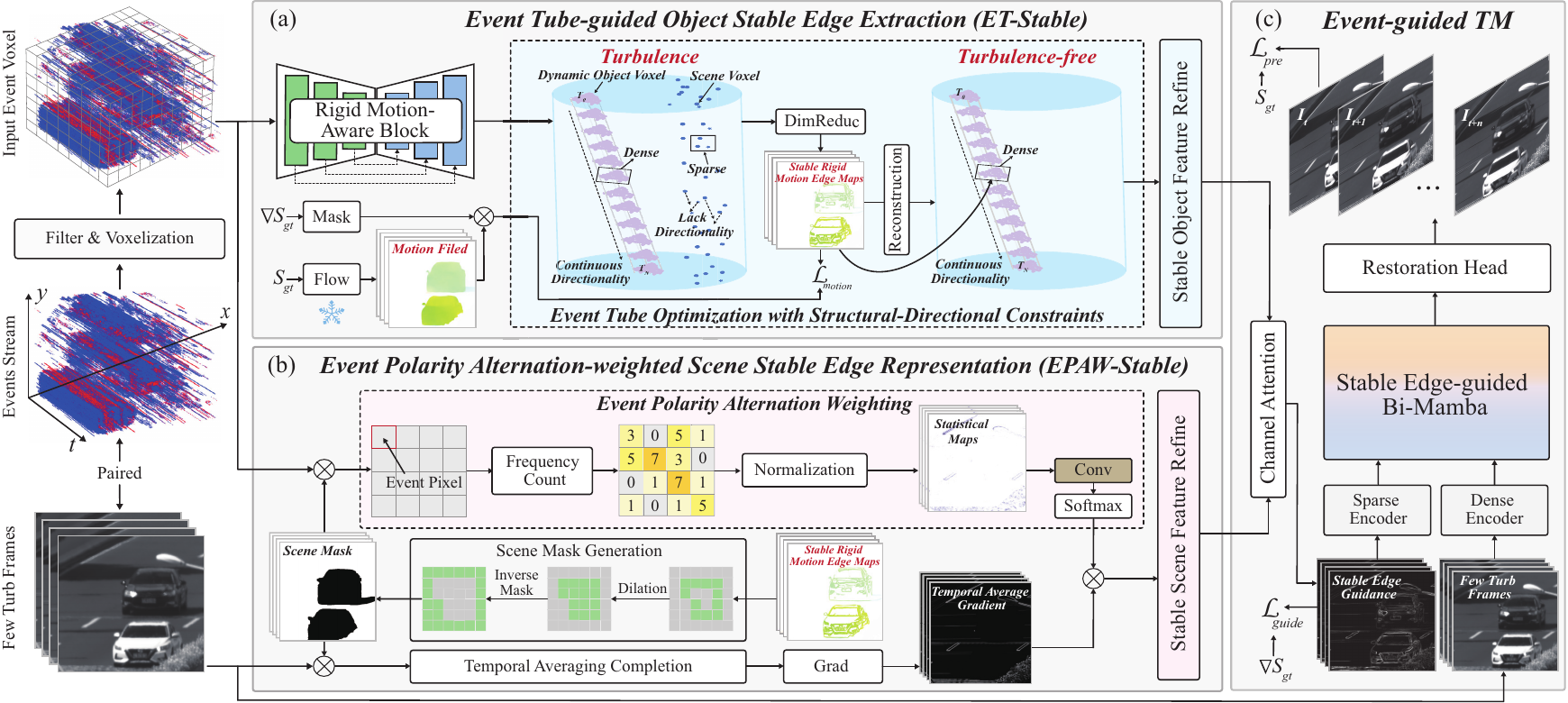}
	\caption{Overall Architecture of the EHETM. (a) ET-Stable stabilizes object motion via the rigid motion-aware block and event tube optimization, leveraging dense spatiotemporal cues to efficiently decouple object motion from turbulence. (b) EPAW-Stable generates masks to isolate turbulence-affected scene regions, and encodes the frequency of event polarity alternation as adaptive weights on temporally averaged gradients to suppress scene distortions. (c) ET-Stable and EPAW-Stable extract stable patterns from turbulence to guide a light-weight Mamba-based video restoration network, enabling high-quality and efficient restoration with few input frames.}
	\label{fig:supp-5}
\end{figure*}

\subsection{Dynamic Object Events: Turb vs. No Turb} 
\label{sec:Dynamic}
From both intuitive and theoretical perspectives, we analyze the response differences between the object and turbulence in events, and reveal how turbulence affects event responses of the object. Event generation process \cite{paredes2021back} is described by:
\begin{equation}
	\setlength{\abovedisplayskip}{5pt}
	\setlength{\belowdisplayskip}{5pt}
	\log I(x, y, t) - \log I(x, y, t - \Delta t) = pC,
	\label{eq:possion-0}
\end{equation}
where $\log I(x, y, t)$ is the logarithmic brightness, $\Delta t$ is the interval, $p \in \{-1, 1\}$ is the polarity, and $C$ is the threshold. 

The spatial displacement of a dynamic object generates continuous brightness changes, forming a coherent event stream, as shown in Fig.~\ref{fig:supp-4} (a). Without turbulence, object events exhibit strong spatiotemporal consistency, marked by well-structured spatial patterns and sustained temporal continuity. Within a short time window (e.g., $40\text{ms}$), this forms a spatiotemporal coherent 3D event tube. This phenomenon benefits from the superiority of events to model continuous motion. It also holds under long-range observations, where small angular velocities ensure the object's geometry and motion state remain nearly unchanged over short periods \cite{gallego2017accurate}. Furthermore, we observe that the event tube can be effectively represented by its 2D motion fields, providing a compact and informative description of objects.

We adopt the notion of ``particle'' from Lagrangian particle tracking~\cite{ali2007lagrangian}, where each particle is represented by a 3D coordinate $(\boldsymbol{x},t)$. Let the trajectory of a particle located at $(\boldsymbol{x}_0,t_0)$ over a temporal window of length $\Delta>0$ be given by a function $T_{\boldsymbol{x}_0}^{t_0}:[t_0-\Delta,t_0+\Delta]\to\mathbb{R}^2$ which maps time to its spatial position, e.g., $T_{\boldsymbol{x}_0}^{t_0}(t_0)=\boldsymbol{x}_0$. The trajectory can be written as the point set $\{(T_{\boldsymbol{x}_0}^{t_0}(t),t):{t\in[t_0-\Delta,t_0+\Delta]\}}$. When $\Delta$ is small, the trajectory can be approximated by the first-order Taylor expansion at $t_0$:
\begin{align}
	T_{\boldsymbol{x}_0}^{t_0}(t)
	&=T_{\boldsymbol{x}_0}^{t_0}(t_0)+(t-t_0)\cdot {T_{\boldsymbol{x}_0}^{t_0}}'(t_0)+o(|t-t_0|) \nonumber\\
	&= \boldsymbol{x}_0+(t-t_0)\cdot {T_{\boldsymbol{x}_0}^{t_0}}'(t_0)+o(\Delta) \nonumber\\
	&\approx \boldsymbol{x}_0+(t-t_0)\cdot {T_{\boldsymbol{x}_0}^{t_0}}'(t_0) \label{clean-trajectory},
\end{align}
where ${T_{\boldsymbol{x}_0}^{t_0}}'(t_0)\in\mathbb{R}^2$ is the first-order derivative. Hence the trajectory is fully determined by this motion vector.

Under turbulence, as shown in Fig.~\ref{fig:supp-4} (b), the spatial structure is distorted. The observed trajectory become
\begin{equation}
	\tilde{T}_{\boldsymbol{x}_0}^{t_0}(t)={T}_{\boldsymbol{x}_0}^{t_0}(t)+\delta(t),
	\label{eq:possion-3}
\end{equation}
where $\delta:\mathbb{R}\to\mathbb{R}^2$ is turbulence and cannot be approximated linearly in the given time interval $\Delta$ due to the short coherent time of turbulence. Now consider the trajectories of all particles on a uniform $h\times w$ grid at time $t_0$:
\begin{equation}
	T^{t_0}=\left\{T^{t_0}_{\boldsymbol{x}}:\boldsymbol{x}\in[h]\times[w]\right\},
	\label{eq:possion-4}
\end{equation}
\begin{equation}
	\tilde{T}^{t_0}=\left\{T^{t_0}_{\boldsymbol{x}}+\delta_{\boldsymbol{x}}:\boldsymbol{x}\in[h]\times[w]\right\},
	\label{eq:possion-5}
\end{equation}
where $[n]=\{1,2,\dots, n\}$ for any positive integer $n$, $T^{t_0}$ is the set of clean trajectories and $\tilde{T}^{t_0}$ is the turbulent version.

Since each clean trajectory is linearized by Eq.~\eqref{clean-trajectory}, the entire set $T^{t_0}$ can be compactly represented by a tensor ${T^{t_0}}'(t_0)\in\mathbb{R}^{h\times w\times 2}$. In contrast, $\tilde{T}^{t_0}$ lacks such a low-dimensional representation. This contrast highlights the intrinsic reducibility of clean 3D tubes, which provides a strong structural prior for recovering turbulence-free trajectories from distorted observations.

\section{Event-guided Turbulence Mitigation}
\label{sec:Method}
\crefname{table}{Table}{Tables} 
\bibliographystyle{unsrt}

\subsection{Event Tube-guided Object Edge Extraction} Dynamic objects generate spatiotemporally coherent event tubes, offering dense temporal cues absent in frame-based methods. Building on this finding, we design the ET-Stable module, which exploits event spatiotemporal continuity to decouple object motion from turbulence and generate stable guidance of objects for TM, as illustrated in Fig.~\ref{fig:supp-5} (a).

\noindent\textbf{Rigid Motion-Aware Block.} While conceptually intuitive, directly modeling distinctive event tubes of dynamic objects is challenging due to the randomness of turbulence. Consequently, we propose a lightweight rigid motion-aware block (RMAB) integrating 3D residual convolutions (3D-ResConvs) with 3D channel attention units (3D-CAUs) to extract spatiotemporally consistent dense motion features from events. Specifically, multi-scale learning strategy is employed on 3D-ResConvs to capture spatiotemporal patterns across scales, while 3D-CAUs adaptively enhance and decode deep features along spatiotemporal dimensions. Skip connections enable effective multi-scale fusion following classic U-Net \cite{ronneberger2015u}. To balance capability and efficiency, we employ only three layers of each 3D-ResConvs and 3D-CAUs for lightweight spatiotemporal feature extraction.

\begin{table*}[t]
	\setlength{\abovecaptionskip}{2pt}
	\setlength{\belowcaptionskip}{-2pt}
	\centering
	\renewcommand{\arraystretch}{1.35}
	\caption{Quantitative comparisons on CTTH, LATH and UDET datasets. The best results are in \textbf{bold} and the second-best are \underline{underlined}.}
	\fontsize{5.8}{5.8}\selectfont
	\begin{tabular}{@{}l*{10}{c}@{}}
		\toprule[1.0pt]
		\multirow{3}{*}{\textbf{Methods}} & \multirow{3}{*}{\textbf{Task / Input Type}} 
		& \multicolumn{3}{c}{\textbf{CTTH / UDET Datasets}} 
		& \textbf{LATH Dataset} 
		& \multicolumn{5}{c}{\textbf{Efficiency (512 × 512 Patches)}} \\
		\cmidrule(lr){3-5}
		\cmidrule(lr){6-6}
		\cmidrule(lr){7-11}
		& & PSNR$\uparrow$ & SSIM$\uparrow$ & LPIPS$\downarrow$ & NIQE$\downarrow$
		& \#Params (M)$\downarrow$ & FPS$\uparrow$ & FLOPs (G)$\downarrow$ & Latency (ms)$\downarrow$ & Data Size (\%)$\downarrow$ \\
		\midrule[1.0pt]
		TurbNet \cite{mao2022single} & TM / Single-frame      
		& 28.17 / 24.12 & 0.8315 / 0.7784 & 0.2210 / 0.2679 & 16.15 & 26.6  & 5.52 & 686.7 & --   & \textbf{2.5}   \\
		VRT \cite{liang2024vrt} & VR / Multi-frame          
		& 32.55 / 26.58 & 0.8986 / 0.8139 & 0.1613 / 0.2334 & 14.37 & 18.3  & 0.31 & 2509.8 & 600    & 40.0  \\
		TSRWGAN \cite{jin2021neutralizing} & TM / Multi-frame 
		& 30.19 / 26.72 & 0.8569 / 0.8027 & 0.1812 / 0.2358 & 13.87 & 46.3  & 1.33 & 1280.3 & \underline{280}   & 37.5  \\
		TMT \cite{zhang2024imaging} & TM / Multi-frame         
		& 32.56 / 27.73 & 0.9046 / 0.8449 & 0.1743 / 0.1975 & 13.09 & 26.1  & 0.11 & 463.9 & 760    & 50.0 \\
		DATUM \cite{zhang2024spatio} & TM / Multi-frame         
		& \underline{34.89} / 28.85 & \underline{0.9340} / \underline{0.8688} & 0.1448 / \underline{0.1565} & \underline{10.87} & \underline{5.8}   & 12.1 & 191.5 & 1560   & 100.0 \\
		MambaTM \cite{zhang2025learning} & TM / Multi-frame  
		& 34.38 / \underline{28.91} & 0.9219 / 0.8610 & \underline{0.1290} / 0.1678 & 11.49 & 6.9   & \underline{23.5} & \textbf{62.3} & 760    & 75.0  \\
		Ev-DeblurVSR \cite{kai2025event} & Deblur / Hybrid 
		& 31.82 / 25.11 & 0.9188 / 0.7961 & 0.1897 / 0.2376 & 15.48 & 8.3   & 2.73 & 448.8 & 560    & 68.1  \\
		EGVSR \cite{lu2023learning} & VSR / Hybrid            
		& 30.15 / 24.27 & 0.8838 / 0.7882 & 0.2189 / 0.2539 & 14.19 & 10.2 & 1.06 & 273.8 & 360    & 33.6  \\
		\textbf{EHETM [Ours]} & \textbf{TM / Hybrid}                          
		& \textbf{35.17} / \textbf{29.33} & \textbf{0.9425} / \textbf{0.8717} & \textbf{0.1281} / \textbf{0.1517} & \textbf{10.12} & \textbf{5.6}   & \textbf{29.5} & \underline{181.7} & \textbf{160} & \underline{22.7} \\
		\bottomrule[1.0pt]
	\end{tabular}
	\label{tab:comparison}
	\vspace{-3pt} 
	\parbox{\textwidth}{%
		\scriptsize
		\textit{Note.} 
		The number of inference frames for each multi-frame method follows the configuration reported in their original paper. FLOPs are computed by averaging the total operations over 100 inference frames. Data Size indicates the total data overhead, including both image and event inputs.%
	}
\end{table*}

\noindent\textbf{Event Tube Optimization.} Dimensionality-reduced event tube optimization (ETO) aims to constrain object motion by exploiting the high-temporal-resolution information of events to project the 3D spatiotemporal volume onto a 2D space, effectively decoupling object motion from turbulence while also improving the computational efficiency. Building upon the dense spatiotemporal cues extracted by the rigid motion–aware block, this projection enforces motion consistency along event tubes. We find that $T^{t_0}$ in Eq.~\eqref{eq:possion-4} can be represented by the clean optical flow, which can model the ideal object motion. Accordingly, we use a pre-trained RAFT model \cite{teed2020raft} to derive clean motion fields that serve as references for optimizing consistent event tubes, enabling the model to preserve true object structures while effectively stabilizing the turbulence. Besides, we employ clean gradient masks to clean motion fields, yielding stable edge-aware motion maps. This manner enhances the event voxel alignment and accelerate convergence by reducing redundancy in flat regions. The resulting stable edge motion fields are reconstructed into structured 3D representations and refined through an encoder, yielding stable dynamic object edge guidance with structural-directional constraints.

\subsection{Event-weighted Scene Edge Representation} Building on ET-Stable for dynamic object stabilization, we further design the EPAW-Stable to handle turbulence in scenes, as shown in Fig.~\ref{fig:supp-5} (b). EPAW-Stable leverages stable object edge maps from ET-Stable as a foundation to generate scene masks, which are applied to both event voxels and frame sequences to isolate areas affected solely by turbulence. Then, we perform temporal averaging completion \cite{saha2024turb} and gradient computation, resulting in initial gradient maps. Simultaneously, event polarity alternation weighting (EPAW) encodes counts of event polarity alternations as adaptive weights for initial gradient maps, sharpening the edges of scenes, while an encoder further refines features to ensure consistent representations. ET-Stable and EPAW-Stable form a complementary framework that exploits high-speed events to stabilize both dynamic objects and scenes, accelerating TM while enhancing restoration quality.

\subsection{Event-guided Video Restoration} Sequence models with linear complexity, such as Mamba \cite{gu2024mamba, zhang2025learning, guo2024mambair, vim}, have shown remarkable efficiency in TM and other vision tasks. Thus, we propose a stable edge-guided Bi-Mamba, as shown in Fig.~\ref{fig:supp-5} (c), which, together with few frames, leverages fused outputs from ET-Stable and EPAW-Stable as full-scene edge guidance. A dual-encoder extracts sparse edge and dense texture features, with shallow features processed via local Hilbert scanning \cite{wu2024rainmamba} before deep feature extraction. Edge priors guide feature evolution, achieving faster and higher-quality restoration. Network details are provided in the supplementary material.


\begin{figure}[t]
	\setlength{\abovecaptionskip}{5pt}
	\setlength{\belowcaptionskip}{-3pt}
	\centering
	\includegraphics[width=\linewidth]{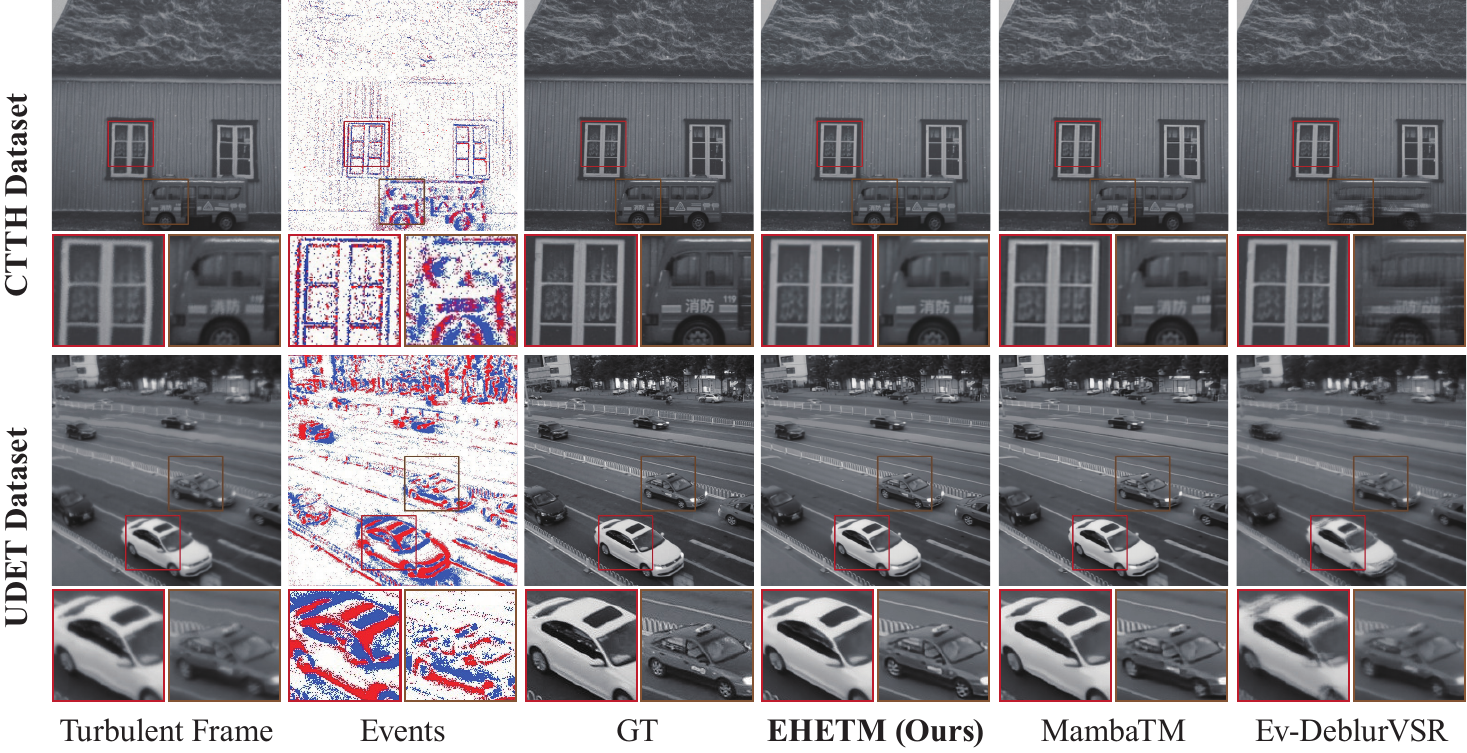}
	\caption{Visual comparisons on CTTH and UDET datasets. Our EHETM yields more visual pleasing results, especially around regions with moving objects.}
	\label{fig:supp-7}
\end{figure}

\subsection{Loss Functions} 
The overall loss function of the EHETM is defined as
\begin{equation}
	\setlength{\abovedisplayskip}{5pt}
	\setlength{\belowdisplayskip}{5pt}
	\mathcal{L}_{\mathrm{total}} = \mathcal{L}_{\mathrm{motion}} + \mathcal{L}_{\mathrm{guide}} + \mathcal{L}_{\mathrm{pre}}.
	\label{eq:possion-6}
\end{equation}
Motion consistency loss $\mathcal{L}_{\mathrm{motion}} = \mathcal{L}_{\mathrm{char}}(\hat{\mathbf{M}}, \mathbf{M})$ ensures the extraction of stable edge motion fields of objects, where $\hat{\mathbf{M}}$ and $\mathbf{M}$ denote the predicted and RAFT-generated motion fields, respectively. Prior loss $\mathcal{L}_{\mathrm{guide}} = \mathcal{L}_{\mathrm{char}}(\hat{\mathbf{G}}, \mathbf{G})$ supervises the ET-Stable and EPAW-Stable to generate stable priors, where $\hat{\mathbf{G}}$ and $\mathbf{G}$ represent the predicted and clean gradients. Restoration loss $\mathcal{L}_{\mathrm{pre}} = \mathcal{L}_{char}(\hat{\mathbf{J}}, \mathbf{J}) + \alpha_{p}\mathcal{L}_{p}(\hat{\mathbf{J}}, \mathbf{J})$ ensures the quality of outputs, where $\hat{\mathbf{J}}$ and $\mathbf{J}$ denote restoration results and GTs. $\mathcal{L}_{\mathrm{char}}$ represents the Charbonnier loss \cite{charbonnier1997deterministic} and $\mathcal{L}_{p}$ represents the perceptual loss \cite{zhang2018unreasonable}, and the weight coefficient $\alpha_{p}$ is set to 0.3.

\section{Experiments}
\label{sec:Experiments}

\begin{figure*}[t]
	\setlength{\abovecaptionskip}{2pt}
	\setlength{\belowcaptionskip}{-2pt}
	\centering
	\includegraphics[width=\linewidth]{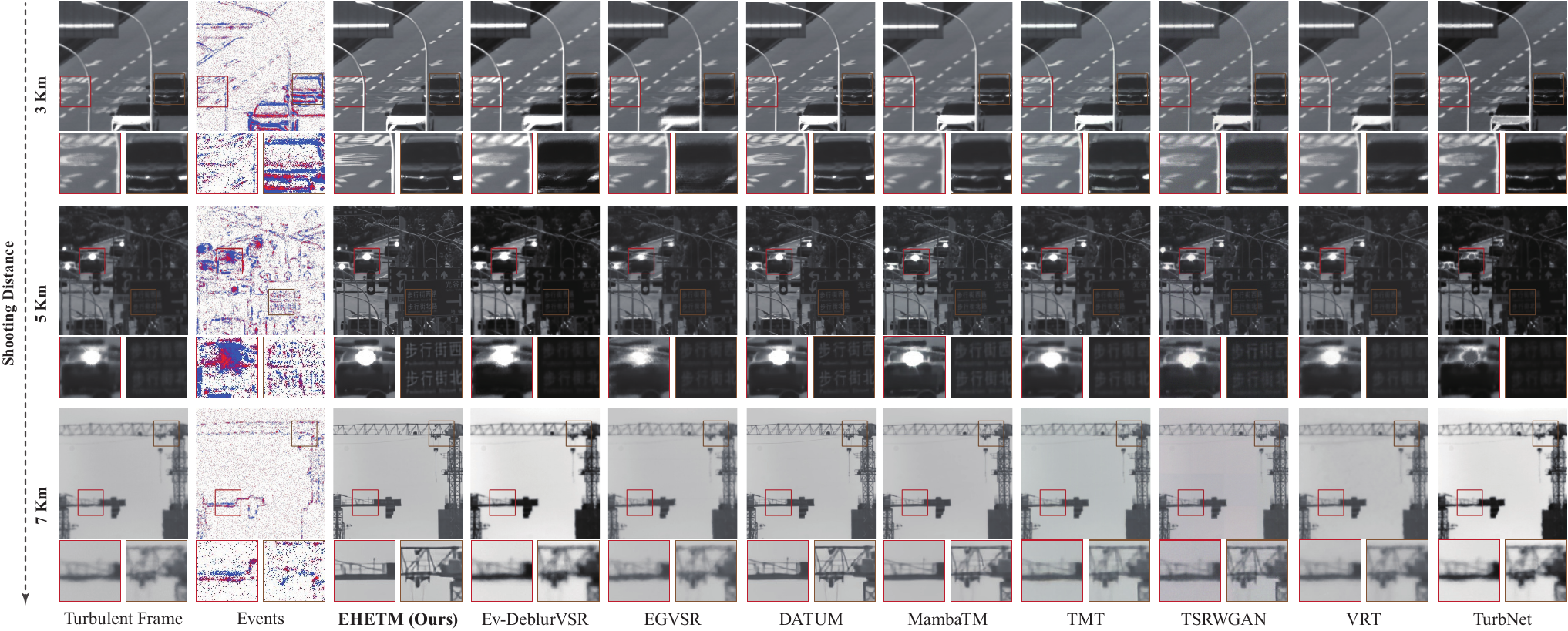}
	\caption{Visual comparisons on LATH dataset. EHETM produces more satisfactory visual results and achieves outstanding generalization under different real-world scenes. The number of inference frames from EHETM to VRT are as follows: \textbf{8}, 15, 10, 40, 20, 20, 15, 16.}
	\label{fig:supp-8}
\end{figure*}

\subsection{Experimental Settings} 
\noindent\textbf{Training Settings.} Our network is implemented in PyTorch \cite{paszke2017automatic} and optimized with Adam \cite{kingma2014adam} using an initial learning rate of $2\times10^{-4}$, scheduled by cosine annealing. Training is conducted on 8 NVIDIA RTX 3090 GPUs, and the detailed training scheme is provided in the supplementary material.

\noindent\textbf{Comparison Methods.} We select eight representative competing methods, including TurbNet \cite{mao2022single}, VRT \cite{liang2024vrt}, TSRWGAN \cite{jin2021neutralizing}, TMT \cite{zhang2024imaging}, DATUM \cite{zhang2024spatio}, MambaTM \cite{zhang2025learning}, Ev-DeblurVSR \cite{kai2025event}, and EGVSR \cite{lu2023learning}. The last two are hybrid-based deblurring and video super-resolution (VSR) methods, while the others are frame-based TM or video restoration (VR) approaches. For the fair comparison, all methods use their released pretrained weights and are additionally re-trained on our CTTH dataset.

\noindent\textbf{Datasets.} To comprehensively evaluate our method, we further adopt the UA-DETRAC dataset \cite{wen2020ua} in addition to our CTTH and LATH datasets. Following \cite{zhang2024spatio}, different levels of turbulence are simulated, and the corresponding events are synthesized with the v2e simulator \cite{hu2021v2e}, yielding the UA-DETRAC-EvTurb (UDET) dataset. 

\noindent\textbf{Evaluation Metrics.} For CTTH and UDET datasets, we evaluate the restoration performance using PSNR \cite{wang2006modern}, SSIM \cite{wang2004image}, and LPIPS \cite{zhang2018unreasonable}, and report efficiency including \#Params, FPS, data size, FLOPs and the system latency. For LATH dataset, which lacks GTs, we adopt natural image quality evaluator (NIQE) \cite{mittal2012making} to assess visual quality of restored images in a no-reference manner.

\subsection{Quantitative Evaluation} 
Table~\ref{tab:comparison} demonstrates that all multi-frame methods outperform the single-frame one, highlighting the merits of temporal cues in mitigating turbulence distortions. Notably, guided by high-temporal-resolution events, our EHETM achieves the best restoration (+0.38~PSNR, +0.0057~SSIM, $-$0.0108~LPIPS, $-$0.75~NIQE vs.~the second best: DATUM \cite{zhang2024spatio}) by suppressing turbulence-induced distortions and stabilizing object edges. Furthermore, with only 5.6~M parameters, 181.7 FLOPs and 5 input frames with corresponding events (22.7\% data size), EHETM attains the highest 29.5 FPS and lowest 160 ms system latency, highlighting its efficiency and practical applicability.

\begin{table}[t]
	\setlength{\abovecaptionskip}{1pt}
	\setlength{\belowcaptionskip}{-1pt}
	\centering
	\renewcommand{\arraystretch}{1.0}
	\caption{Ablation studies of EHETM}
	\fontsize{9}{9}\selectfont
	\begin{tabular}{ccc!{\vrule width 1.0pt}ccc}
		\toprule[1.0pt]
		RMAB & ETO & EPAW & PSNR$\uparrow$ & SSIM$\uparrow$ & LPIPS$\downarrow$ \\ 
		\midrule[1.0pt]
		$\checkmark$ &              &              &31.13 &0.8962 &0.2261       \\
		& $\checkmark$ &              &33.85 &0.9241 &0.1736       \\
		$\checkmark$ & $\checkmark$ &              &34.19 &0.9309 &0.1563       \\ 
		$\checkmark$ & $\checkmark$ & $\checkmark$ &\textbf{35.17} &\textbf{0.9425} &\textbf{0.1281}       \\ 
		\bottomrule[1.0pt]
	\end{tabular}
	\label{tab:Ablation}
\end{table}

\subsection{Qualitative Evaluation} 
\noindent\textbf{Evaluation on CTTH and UDET Datasets.} Fig.~\ref{fig:supp-7} depicts a visual comparison between EHETM and two typical methods. EHETM produces sharper edges and more coherent structures, particularly around dynamic objects. In contrast, MambaTM \cite{zhang2025learning} and Ev-DeblurVSR \cite{kai2025event} exhibit blurry edges and residual distortions. These results highlight that using temporal cues from high-speed events is crucial to TM, leading to higher-quality restoration. 

\noindent\textbf{Evaluation on LATH Dataset.} We compare EHETM with eight typical methods for generalization evaluation, as shown in Fig.~\ref{fig:supp-8}. TurbNet \cite{mao2022single} struggles with distortions as it lacks of temporal modeling for TM. VRT \cite{liang2024vrt}, TSRWGAN \cite{jin2021neutralizing}, and TMT \cite{zhang2024imaging} partially reduce turbulence effects, but some blur and noise remain unsolved. Ev-DeblurVSR \cite{kai2025event} alleviates blur at the cost of degraded contrast, and EGVSR \cite{lu2023learning} introduces unexpected artifacts. While DATUM \cite{zhang2024spatio} and MambaTM \cite{zhang2025learning} perform reasonably well, severe aliasing between object motion and turbulence still causes deviations. In contrast, our EHETM consistently produces superior results across different scenes, suppressing turbulence distortions while stabilizing dynamic objects.

\subsection{Ablation Study and Discussion} 
\begin{figure}[t]
	\setlength{\abovecaptionskip}{5pt}
	\setlength{\belowcaptionskip}{-5pt}
	\centering
	\includegraphics[width=\linewidth]{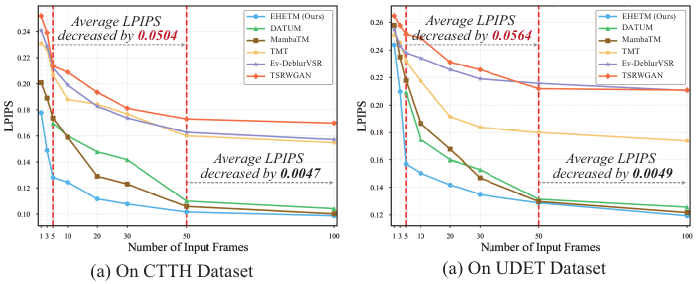}
	\caption{Impact of the inference data size. With only few frames and events, our method demonstrates comparable or even higher quality restoration compared to other multi-frame methods.}
	\label{fig:supp-12}
\end{figure}

\noindent\textbf{Impact of the Inference Data Size.} We investigate the impact of inference data size on TM, as shown in Fig.~\ref{fig:supp-12}. With very limited inputs (1 to 5 frames), EHETM excels by using the high-speed events to capture the zero-mean characteristics of turbulence. As data size grows (5 to 50 frames), all methods improve substantially (average LPIPS reduction $>$ 0.5), yet EHETM retains a clear advantage due to finer-grained temporal cues. For larger data size (50 to 100 frames), gains plateau (average LPIPS reduction $<$ 0.05) while computational costs rise sharply, highlighting the importance of efficiently mining temporal information for TM.

\begin{figure}[t]
	\setlength{\abovecaptionskip}{5pt}
	\setlength{\belowcaptionskip}{-5pt}
	\centering
	\includegraphics[width=\linewidth]{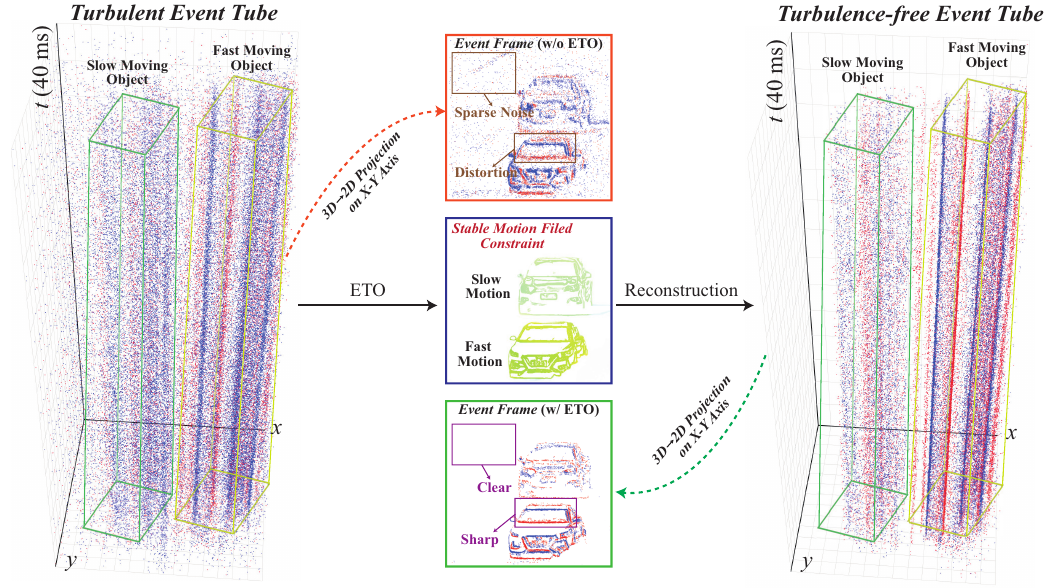}
	\caption{Effectiveness of ETO. It extracts clean event tubes triggered by objects with various motions from turbulent events.}
	\label{fig:supp-9}
\end{figure}

\noindent\textbf{Effectiveness of Event Tube Optimization.} As shown in Fig.~\ref{fig:supp-9}, using 40 ms accumulated event voxels, we find that w/o ETO, turbulence-corrupted event tubes produce sparse noise and distorted edges when projected to the event frame. By applying 3D-to-2D constraint, ETO suppresses these distortions while maintaining sharp edges and scene clarity, yielding spatiotemporally consistent event tubes. It is noteworthy that ETO effectively constrains objects even when their motion speeds differ. Table~\ref{tab:Ablation} further confirms the effectiveness of ETO, and when integrated with RMAB, it enables effective object edge stabilization and significantly contributes to high-quality restoration.

\begin{figure}[t]
	\setlength{\abovecaptionskip}{5pt}
	\setlength{\belowcaptionskip}{-5pt}
	\centering
	\includegraphics[width=\linewidth]{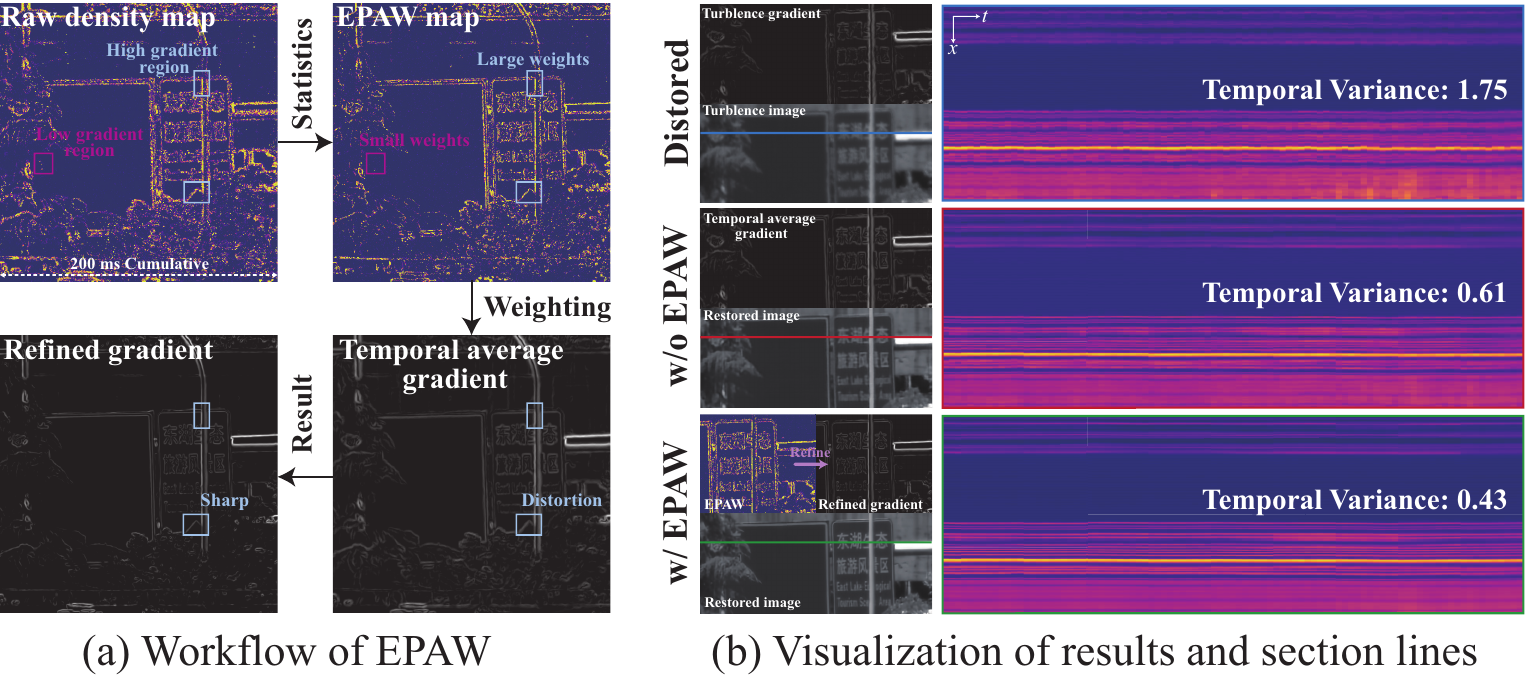}
	\caption{Effectiveness of EPAW. (a) How does EPAW work. (b) Visualization of selected section lines and the variance computed along the \textit{t} axis.}
	\label{fig:supp-10}
\end{figure}

\noindent\textbf{Effectiveness of Event Polarity Alternation Weighting.} EPAW refines distorted edges within the temporal average gradient to guide accurate scene restoration. In Fig.~\ref{fig:supp-10} (a), we illustrate the workflow of EPAW. By computing the EPAW map, sharp structures in high-gradient regions are assigned greater weights, thereby optimizing distorted gradients and producing sharper edge representations. As shown by the selected section lines in Fig.~\ref{fig:supp-10} (b), w/ EPAW produces smoother results, and the temporal variance along these lines quantitatively verifies the improvement in stability. Moreover, Table~\ref{tab:Ablation} verifies the improvement in overall restoration brought by EPAW. In summary, w/ EPAW enables the model to learn sharp edges during end-to-end training, resulting in higher-quality scene restoration.

\noindent\textbf{Limitation and Future Work.} EHETM exhibits high efficiency but faces challenges in ultra-long-range scenes (e.g. $>$ 30 km) and under severe platform motion (e.g. fast object tracking). Ultra-long-range atmospheric sparsifies event responses, lacking information for reliable restoration. Severe platform motion entangles turbulence, scenes and objects, making motion disentanglement extremely challenging. In future, we plan to introduce inertial measurement unit and ultra-telephoto optics to overcome these problems.

\section{Conclusion}
\label{sec:Conclusion}
In this work, we present an event-guided high-quality and efficient TM method. By mining fine-grained motion cues from events within a short duration, we overcome the accuracy–efficiency bottleneck in frames-based methods. With experimental and theoretical analysis, we reveal distinct spatiotemporal characteristics of turbulent events—scenes exhibit polarity alternation while objects form event tubes. Based on these insights, we leverage polarity-weighted gradients for scene refinement and event-tube priors for motion decoupling. Moreover, for comprehensive demonstration, we construct two real-world event-frame paired turbulence datasets: CTTH and LATH. Our method achieves superior restoration quality and efficiency over SOTAs, and we believe it can bring new perspectives to the community. 

\clearpage
\newpage

\noindent
\textbf{Acknowledgments.}
This work was supported by the National Natural Science Foundation of China under Grant U24B20139 and 62371203, the Open Fund of the State Key Laboratory of Dynamic Optical Imaging and Measurement E41302A1L1, the Hubei Provincial Natural Science Foundation under Grant JCZRQT202600099.

{
	\small
	\bibliographystyle{ieeenat_fullname}
	\bibliography{main}

\begin{thebibliography}{10}

\bibitem{jin2021neutralizing}
Darui Jin, Ying Chen, Yi~Lu, Junzhang Chen, Peng Wang, Zichao Liu, Sheng Guo,
  and Xiangzhi Bai.
\newblock Neutralizing the impact of atmospheric turbulence on complex scene
  imaging via deep learning.
\newblock {\em Nature Machine Intelligence}, 3(10):876--884, 2021.

\bibitem{guo2024direct}
Yuduo Guo, Yuhan Hao, Sen Wan, Hao Zhang, Laiyu Zhu, Yi~Zhang, Jiamin Wu,
  Qionghai Dai, and Lu~Fang.
\newblock Direct observation of atmospheric turbulence with a video-rate
  wide-field wavefront sensor.
\newblock {\em Nature Photonics}, 18(9):935--943, 2024.

\bibitem{hill2025deep}
Paul Hill, Nantheera Anantrasirichai, Alin Achim, and David Bull.
\newblock Deep learning techniques for atmospheric turbulence removal: a
  review.
\newblock {\em Artificial Intelligence Review}, 58(4):101, 2025.

\bibitem{anantrasirichai2013atmospheric}
Nantheera Anantrasirichai, Alin Achim, Nick~G Kingsbury, and David~R Bull.
\newblock Atmospheric turbulence mitigation using complex wavelet-based fusion.
\newblock {\em IEEE Transactions on Image Processing}, 22(6):2398--2408, 2013.

\bibitem{chan2022tilt}
Stanley~H Chan.
\newblock Tilt-then-blur or blur-then-tilt? clarifying the atmospheric
  turbulence model.
\newblock {\em IEEE Signal Processing Letters}, 29:1833--1837, 2022.

\bibitem{cornett2023expanding}
David Cornett, Joel Brogan, Nell Barber, Deniz Aykac, Seth Baird, Nicholas
  Burchfield, Carl Dukes, Andrew Duncan, Regina Ferrell, Jim Goddard, et~al.
\newblock Expanding accurate person recognition to new altitudes and ranges:
  The briar dataset.
\newblock In {\em Proceedings of the IEEE/CVF Winter Conference on Applications
  of Computer Vision}, pages 593--602, 2023.

\bibitem{zhang2024source}
Xingguang Zhang and Chih-Hsien Chou.
\newblock Source-free domain adaptation for video object detection under
  adverse image conditions.
\newblock In {\em Proceedings of the IEEE/CVF Conference on Computer Vision and
  Pattern Recognition}, pages 5010--5019, 2024.

\bibitem{zhang2024spatio}
Xingguang Zhang, Nicholas Chimitt, Yiheng Chi, Zhiyuan Mao, and Stanley~H Chan.
\newblock Spatio-temporal turbulence mitigation: A translational perspective.
\newblock In {\em Proceedings of the IEEE/CVF conference on computer vision and
  pattern recognition}, pages 2889--2899, 2024.

\bibitem{hirsch2010efficient}
Michael Hirsch, Suvrit Sra, Bernhard Sch{\"o}lkopf, and Stefan Harmeling.
\newblock Efficient filter flow for space-variant multiframe blind
  deconvolution.
\newblock In {\em 2010 IEEE Computer Society Conference on Computer Vision and
  Pattern Recognition}, pages 607--614. IEEE, 2010.

\bibitem{patel2019adaptive}
Akshay Patel, Dippal Israni, Nerella Arun~Mani Kumar, and Chintan Bhatt.
\newblock An adaptive image registration technique to remove atmospheric
  turbulence.
\newblock {\em Statistics, Optimization \& Information Computing},
  7(2):439--446, 2019.

\bibitem{caliskan2014atmospheric}
Tufan Caliskan and Nafiz Arica.
\newblock Atmospheric turbulence mitigation using optical flow.
\newblock In {\em 2014 22nd International Conference on Pattern Recognition},
  pages 883--888. Ieee, 2014.

\bibitem{fazlali2022atmospheric}
Hamidreza Fazlali, Shahram Shirani, Michael BradforSd, and Thia Kirubarajan.
\newblock Atmospheric turbulence removal in long-range imaging using a
  data-driven-based approach.
\newblock {\em International Journal of Computer Vision}, 130(4):1031--1049,
  2022.

\bibitem{zhu2012removing}
Xiang Zhu and Peyman Milanfar.
\newblock Removing atmospheric turbulence via space-invariant deconvolution.
\newblock {\em IEEE transactions on pattern analysis and machine intelligence},
  35(1):157--170, 2012.

\bibitem{qin2024unsupervised}
Dehao Qin, Ripon~Kumar Saha, Woojeh Chung, Suren Jayasuriya, Jinwei Ye, and
  Nianyi Li.
\newblock Unsupervised moving object segmentation with atmospheric turbulence.
\newblock In {\em European Conference on Computer Vision}, pages 18--37.
  Springer, 2024.

\bibitem{xu2024long}
Shengqi Xu, Run Sun, Yi~Chang, Shuning Cao, Xueyao Xiao, and Luxin Yan.
\newblock Long-range turbulence mitigation: a large-scale dataset and a
  coarse-to-fine framework.
\newblock In {\em European Conference on Computer Vision}, pages 311--329.
  Springer, 2024.

\bibitem{jaiswal2023physics}
Ajay Jaiswal, Xingguang Zhang, Stanley~H Chan, and Zhangyang Wang.
\newblock Physics-driven turbulence image restoration with stochastic
  refinement.
\newblock In {\em Proceedings of the IEEE/CVF international conference on
  computer vision}, pages 12170--12181, 2023.

\bibitem{lau2021atfacegan}
Chun~Pong Lau, Carlos~D Castillo, and Rama Chellappa.
\newblock Atfacegan: Single face semantic aware image restoration and
  recognition from atmospheric turbulence.
\newblock {\em IEEE Transactions on Biometrics, Behavior, and Identity
  Science}, 3(2):240--251, 2021.

\bibitem{mao2022single}
Zhiyuan Mao, Ajay Jaiswal, Zhangyang Wang, and Stanley~H Chan.
\newblock Single frame atmospheric turbulence mitigation: A benchmark study and
  a new physics-inspired transformer model.
\newblock In {\em European conference on computer vision}, pages 430--446.
  Springer, 2022.

\bibitem{mei2023ltt}
Kangfu Mei and Vishal~M Patel.
\newblock Ltt-gan: Looking through turbulence by inverting gans.
\newblock {\em IEEE Journal of Selected Topics in Signal Processing},
  17(3):587--598, 2023.

\bibitem{nair2021confidence}
Nithin~Gopalakrishnan Nair and Vishal~M Patel.
\newblock Confidence guided network for atmospheric turbulence mitigation.
\newblock In {\em 2021 IEEE International Conference on Image Processing
  (ICIP)}, pages 1359--1363. IEEE, 2021.

\bibitem{nair2023ddpm}
Nithin~Gopalakrishnan Nair, Kangfu Mei, and Vishal~M Patel.
\newblock At-ddpm: Restoring faces degraded by atmospheric turbulence using
  denoising diffusion probabilistic models.
\newblock In {\em Proceedings of the IEEE/CVF winter conference on applications
  of computer vision}, pages 3434--3443, 2023.

\bibitem{rai2022removing}
Shyam~Nandan Rai and CV~Jawahar.
\newblock Removing atmospheric turbulence via deep adversarial learning.
\newblock {\em IEEE Transactions on Image Processing}, 31:2633--2646, 2022.

\bibitem{yasarla2022cnn}
Rajeev Yasarla and Vishal~M Patel.
\newblock Cnn-based restoration of a single face image degraded by atmospheric
  turbulence.
\newblock {\em IEEE Transactions on Biometrics, Behavior, and Identity
  Science}, 4(2):222--233, 2022.

\bibitem{lau2021semi}
Chun~Pong Lau, Amit Kumar, and Rama Chellappa.
\newblock Semi-supervised landmark-guided restoration of atmospheric turbulent
  images.
\newblock {\em IEEE Journal of Selected Topics in Signal Processing},
  15(2):204--215, 2021.

\bibitem{chen2019u}
Gongping Chen, Zhisheng Gao, Qiaolu Wang, and Qingqing Luo.
\newblock U-net like deep autoencoders for deblurring atmospheric turbulence.
\newblock {\em Journal of Electronic Imaging}, 28(5):053024--053024, 2019.

\bibitem{yasarla2021learning}
Rajeev Yasarla and Vishal~M Patel.
\newblock Learning to restore images degraded by atmospheric turbulence using
  uncertainty.
\newblock In {\em 2021 IEEE international conference on image processing
  (ICIP)}, pages 1694--1698. IEEE, 2021.

\bibitem{wu2024semi}
Yubo Wu, Kuanhong Cheng, Ting Cao, Dong Zhao, and Junhuai Li.
\newblock Semi-supervised correction model for turbulence-distorted images.
\newblock {\em Optics Express}, 32(12):21160--21174, 2024.

\bibitem{saha2024turb}
Ripon~Kumar Saha, Dehao Qin, Nianyi Li, Jinwei Ye, and Suren Jayasuriya.
\newblock Turb-seg-res: A segment-then-restore pipeline for dynamic videos with
  atmospheric turbulence.
\newblock In {\em Proceedings of the IEEE/CVF Conference on Computer Vision and
  Pattern Recognition}, pages 25286--25296, 2024.

\bibitem{anantrasirichai2023atmospheric}
Nantheera Anantrasirichai.
\newblock Atmospheric turbulence removal with complex-valued convolutional
  neural network.
\newblock {\em Pattern Recognition Letters}, 171:69--75, 2023.

\bibitem{wang2023revelation}
Yadong Wang, Darui Jin, Junzhang Chen, and Xiangzhi Bai.
\newblock Revelation of hidden 2d atmospheric turbulence strength fields from
  turbulence effects in infrared imaging.
\newblock {\em Nature Computational Science}, 3(8):687--699, 2023.

\bibitem{gao2019atmospheric}
Jing Gao, Nantheera Anantrasirichai, and David Bull.
\newblock Atmospheric turbulence removal using convolutional neural network.
\newblock {\em arXiv preprint arXiv:1912.11350}, 2019.

\bibitem{chak2018subsampled}
Wai~Ho Chak, Chun~Pong Lau, and Lok~Ming Lui.
\newblock Subsampled turbulence removal network.
\newblock {\em arXiv preprint arXiv:1807.04418}, 2018.

\bibitem{xia2025planet}
Yifei Xia, Chu Zhou, Chengxuan Zhu, Chao Xu, and Boxin Shi.
\newblock Planet: Learning to mitigate atmospheric turbulence in planetary
  images.
\newblock In {\em Proceedings of the AAAI Conference on Artificial
  Intelligence}, volume~39, pages 8584--8592, 2025.

\bibitem{feng2023turbugan}
Brandon~Y Feng, Mingyang Xie, and Christopher~A Metzler.
\newblock Turbugan: An adversarial learning approach to spatially-varying
  multiframe blind deconvolution with applications to imaging through
  turbulence.
\newblock {\em IEEE Journal on Selected Areas in Information Theory},
  3(3):543--556, 2023.

\bibitem{liu2023eaf}
Xiangqing Liu, Gang Li, Zhenyang Zhao, Qi~Cao, Zijun Zhang, Shaoan Yan, Jianbin
  Xie, and Minghua Tang.
\newblock Eaf-wgan: Enhanced alignment fusion-wasserstein generative
  adversarial network for turbulent image restoration.
\newblock {\em IEEE Transactions on Circuits and Systems for Video Technology},
  33(10):5605--5616, 2023.

\bibitem{mao2020image}
Zhiyuan Mao, Nicholas Chimitt, and Stanley~H Chan.
\newblock Image reconstruction of static and dynamic scenes through
  anisoplanatic turbulence.
\newblock {\em IEEE Transactions on Computational Imaging}, 6:1415--1428, 2020.

\bibitem{zhang2024imaging}
Xingguang Zhang, Zhiyuan Mao, Nicholas Chimitt, and Stanley~H Chan.
\newblock Imaging through the atmosphere using turbulence mitigation
  transformer.
\newblock {\em IEEE Transactions on Computational Imaging}, 10:115--128, 2024.

\bibitem{zou2024deturb}
Zhicheng Zou and Nantheera Anantrasirichai.
\newblock Deturb: atmospheric turbulence mitigation with deformable 3d
  convolutions and 3d swin transformers.
\newblock In {\em Proceedings of the Asian Conference on Computer Vision},
  pages 904--921, 2024.

\bibitem{wang2024real}
Xijun Wang, Santiago L{\'o}pez-Tapia, and Aggelos~K Katsaggelos.
\newblock Real-world atmospheric turbulence correction via domain adaptation.
\newblock In {\em 2024 IEEE International Conference on Image Processing
  (ICIP)}, pages 1466--1472. IEEE, 2024.

\bibitem{cai2024temporally}
Haoming Cai, Jingxi Chen, Brandon Feng, Weiyun Jiang, Mingyang Xie, Kevin
  Zhang, Cornelia Fermuller, Yiannis Aloimonos, Ashok Veeraraghavan, and Chris
  Metzler.
\newblock Temporally consistent atmospheric turbulence mitigation with neural
  representations.
\newblock {\em Advances in Neural Information Processing Systems},
  37:44554--44574, 2024.

\bibitem{zhang2025learning}
Xingguang Zhang, Nicholas Chimitt, Xijun Wang, Yu~Yuan, and Stanley~H Chan.
\newblock Learning phase distortion with selective state space models for video
  turbulence mitigation.
\newblock In {\em Proceedings of the Computer Vision and Pattern Recognition
  Conference}, pages 2127--2138, 2025.

\bibitem{zhang2024physical}
Ziran Zhang, Yuhang Tang, Zhigang Wang, Yueting Chen, and Bin Zhao.
\newblock Physical prior guided cooperative learning framework for joint
  turbulence degradation estimation and infrared video restoration.
\newblock {\em arXiv preprint arXiv:2408.04227}, 2024.

\bibitem{li2021unsupervised}
Nianyi Li, Simron Thapa, Cameron Whyte, Albert~W Reed, Suren Jayasuriya, and
  Jinwei Ye.
\newblock Unsupervised non-rigid image distortion removal via grid deformation.
\newblock In {\em Proceedings of the IEEE/CVF International Conference on
  Computer Vision}, pages 2522--2532, 2021.

\bibitem{jiang2023nert}
Weiyun Jiang, Vivek Boominathan, and Ashok Veeraraghavan.
\newblock Nert: Implicit neural representations for unsupervised atmospheric
  turbulence mitigation.
\newblock In {\em Proceedings of the IEEE/CVF Conference on Computer Vision and
  Pattern Recognition}, pages 4236--4243, 2023.

\bibitem{gallego2020event}
Guillermo Gallego, Tobi Delbr{\"u}ck, Garrick Orchard, Chiara Bartolozzi, Brian
  Taba, Andrea Censi, Stefan Leutenegger, Andrew~J Davison, J{\"o}rg Conradt,
  Kostas Daniilidis, et~al.
\newblock Event-based vision: A survey.
\newblock {\em IEEE transactions on pattern analysis and machine intelligence},
  44(1):154--180, 2020.

\bibitem{liu2024seeing}
Haoyue Liu, Shihan Peng, Lin Zhu, Yi~Chang, Hanyu Zhou, and Luxin Yan.
\newblock Seeing motion at nighttime with an event camera.
\newblock In {\em Proceedings of the IEEE/CVF Conference on Computer Vision and
  Pattern Recognition}, pages 25648--25658, 2024.

\bibitem{yu2024eventps}
Bohan Yu, Jieji Ren, Jin Han, Feishi Wang, Jinxiu Liang, and Boxin Shi.
\newblock Eventps: Real-time photometric stereo using an event camera.
\newblock In {\em Proceedings of the IEEE/CVF Conference on Computer Vision and
  Pattern Recognition}, pages 9602--9611, 2024.

\bibitem{xia2024nb}
Yifei Xia, Chu Zhou, Chengxuan Zhu, Minggui Teng, Chao Xu, and Boxin Shi.
\newblock Nb-gtr: narrow-band guided turbulence removal.
\newblock In {\em Proceedings of the IEEE/CVF Conference on Computer Vision and
  Pattern Recognition}, pages 24934--24943, 2024.

\bibitem{liu2025timetracker}
Haoyue Liu, Jinghan Xu, Yi~Chang, Hanyu Zhou, Haozhi Zhao, Lin Wang, and Luxin
  Yan.
\newblock Timetracker: Event-based continuous point tracking for video frame
  interpolation with non-linear motion.
\newblock In {\em Proceedings of the Computer Vision and Pattern Recognition
  Conference}, pages 17649--17659, 2025.

\bibitem{xie2025diffusion}
Xinan Xie, Qing Zhang, and Wei-Shi Zheng.
\newblock Diffusion-based event generation for high-quality image deblurring.
\newblock In {\em Proceedings of the Computer Vision and Pattern Recognition
  Conference}, pages 2194--2203, 2025.

\bibitem{xiao2025event}
Zeyu Xiao and Xinchao Wang.
\newblock Event-based video super-resolution via state space models.
\newblock In {\em Proceedings of the Computer Vision and Pattern Recognition
  Conference}, pages 12564--12574, 2025.

\bibitem{duan2024led}
Yuxing Duan.
\newblock Led: A large-scale real-world paired dataset for event camera
  denoising.
\newblock In {\em Proceedings of the IEEE/CVF Conference on Computer Vision and
  Pattern Recognition}, pages 25637--25647, 2024.

\bibitem{zhou2024bring}
Hanyu Zhou, Yi~Chang, and Zhiwei Shi.
\newblock Bring event into rgb and lidar: Hierarchical visual-motion fusion for
  scene flow.
\newblock In {\em Proceedings of the IEEE/CVF Conference on Computer Vision and
  Pattern Recognition}, pages 26477--26486, 2024.

\bibitem{xiao2024event}
Zeyu Xiao, Dachun Kai, Yueyi Zhang, Zheng-Jun Zha, Xiaoyan Sun, and Zhiwei
  Xiong.
\newblock Event-adapted video super-resolution.
\newblock In {\em European Conference on Computer Vision}, pages 217--235.
  Springer, 2024.

\bibitem{jing2021turning}
Yongcheng Jing, Yiding Yang, Xinchao Wang, Mingli Song, and Dacheng Tao.
\newblock Turning frequency to resolution: Video super-resolution via event
  cameras.
\newblock In {\em Proceedings of the IEEE/CVF Conference on Computer Vision and
  Pattern Recognition}, pages 7772--7781, 2021.

\bibitem{kai2023video}
Dachun Kai, Yueyi Zhang, and Xiaoyan Sun.
\newblock Video super-resolution via event-driven temporal alignment.
\newblock In {\em 2023 IEEE International Conference on Image Processing
  (ICIP)}, pages 2950--2954. IEEE, 2023.

\bibitem{pan2019bringing}
Liyuan Pan, Cedric Scheerlinck, Xin Yu, Richard Hartley, Miaomiao Liu, and
  Yuchao Dai.
\newblock Bringing a blurry frame alive at high frame-rate with an event
  camera.
\newblock In {\em Proceedings of the IEEE/CVF conference on computer vision and
  pattern recognition}, pages 6820--6829, 2019.

\bibitem{yu2024learning}
Wei Yu, Jianing Li, Shengping Zhang, and Xiangyang Ji.
\newblock Learning scale-aware spatio-temporal implicit representation for
  event-based motion deblurring.
\newblock In {\em Forty-first International Conference on Machine Learning},
  2024.

\bibitem{zhang2023generalizing}
Xiang Zhang, Lei Yu, Wen Yang, Jianzhuang Liu, and Gui-Song Xia.
\newblock Generalizing event-based motion deblurring in real-world scenarios.
\newblock In {\em Proceedings of the IEEE/CVF International Conference on
  Computer Vision}, pages 10734--10744, 2023.

\bibitem{kim2024cmta}
Taewoo Kim, Hoonhee Cho, and Kuk-Jin Yoon.
\newblock Cmta: Cross-modal temporal alignment for event-guided video
  deblurring.
\newblock In {\em European Conference on Computer Vision}, pages 1--19.
  Springer, 2024.

\bibitem{liu2025evturb}
Yixing Liu, Minggui Teng, Yifei Xia, Peiqi Duan, and Boxin Shi.
\newblock Evturb: Event camera guided turbulence removal.
\newblock {\em arXiv preprint arXiv:2508.10582}, 2025.

\bibitem{li2025egtm}
Huanan Li, Rui Fan, Juntao Guan, Weidong Hao, Lai Rui, Tong Wu, Yikai Wang, and
  Lin Gu.
\newblock Egtm: Event-guided efficient turbulence mitigation.
\newblock {\em arXiv preprint arXiv:2509.03808}, 2025.

\bibitem{ug22023}
{UG2+} challenge at {CVPR} 2023: Bridging the gap between computational
  photography and visual recognition.
\newblock \url{http://cvpr2023.ug2challenge.org/dataset23_t2.html}, 2023.

\bibitem{gilles2017open}
J{\'e}r{\^o}me Gilles and Nicholas~B Ferrante.
\newblock Open turbulent image set (otis).
\newblock {\em Pattern Recognition Letters}, 86:38--41, 2017.

\bibitem{wang2021realesrgan}
Xintao Wang, Liangbin Xie, Chao Dong, and Ying Shan.
\newblock Real-esrgan: Training real-world blind super-resolution with pure
  synthetic data.
\newblock In {\em International Conference on Computer Vision Workshops
  (ICCVW)}.

\bibitem{paredes2021back}
Federico Paredes-Vall{\'e}s and Guido~CHE De~Croon.
\newblock Back to event basics: Self-supervised learning of image
  reconstruction for event cameras via photometric constancy.
\newblock In {\em Proceedings of the IEEE/CVF Conference on Computer Vision and
  Pattern Recognition}, pages 3446--3455, 2021.

\bibitem{gallego2017accurate}
Guillermo Gallego and Davide Scaramuzza.
\newblock Accurate angular velocity estimation with an event camera.
\newblock {\em IEEE Robotics and Automation Letters}, 2(2):632--639, 2017.

\bibitem{ali2007lagrangian}
Saad Ali and Mubarak Shah.
\newblock A lagrangian particle dynamics approach for crowd flow segmentation
  and stability analysis.
\newblock In {\em 2007 IEEE conference on computer vision and pattern
  recognition}, pages 1--6. IEEE, 2007.

\bibitem{ronneberger2015u}
Olaf Ronneberger, Philipp Fischer, and Thomas Brox.
\newblock U-net: Convolutional networks for biomedical image segmentation.
\newblock In {\em International Conference on Medical image computing and
  computer-assisted intervention}, pages 234--241. Springer, 2015.

\bibitem{liang2024vrt}
Jingyun Liang, Jiezhang Cao, Yuchen Fan, Kai Zhang, Rakesh Ranjan, Yawei Li,
  Radu Timofte, and Luc Van~Gool.
\newblock Vrt: A video restoration transformer.
\newblock {\em IEEE Transactions on Image Processing}, 33:2171--2182, 2024.

\bibitem{kai2025event}
Dachun Kai, Yueyi Zhang, Jin Wang, Zeyu Xiao, Zhiwei Xiong, and Xiaoyan Sun.
\newblock Event-enhanced blurry video super-resolution.
\newblock In {\em Proceedings of the AAAI Conference on Artificial
  Intelligence}, volume~39, pages 4175--4183, 2025.

\bibitem{lu2023learning}
Yunfan Lu, Zipeng Wang, Minjie Liu, Hongjian Wang, and Lin Wang.
\newblock Learning spatial-temporal implicit neural representations for
  event-guided video super-resolution.
\newblock In {\em Proceedings of the IEEE/CVF Conference on Computer Vision and
  Pattern Recognition}, pages 1557--1567, 2023.

\bibitem{teed2020raft}
Zachary Teed and Jia Deng.
\newblock Raft: Recurrent all-pairs field transforms for optical flow.
\newblock In {\em European conference on computer vision}, pages 402--419.
  Springer, 2020.

\bibitem{gu2024mamba}
Albert Gu and Tri Dao.
\newblock Mamba: Linear-time sequence modeling with selective state spaces.
\newblock In {\em First Conference on Language Modeling}, 2024.

\bibitem{guo2024mambair}
Hang Guo, Jinmin Li, Tao Dai, Zhihao Ouyang, Xudong Ren, and Shu-Tao Xia.
\newblock Mambair: A simple baseline for image restoration with state-space
  model.
\newblock In {\em European conference on computer vision}, pages 222--241.
  Springer, 2024.

\bibitem{vim}
Lianghui Zhu, Bencheng Liao, Qian Zhang, Xinlong Wang, Wenyu Liu, and Xinggang
  Wang.
\newblock Vision mamba: Efficient visual representation learning with
  bidirectional state space model.
\newblock {\em arXiv preprint arXiv:2401.09417}, 2024.

\bibitem{wu2024rainmamba}
Hongtao Wu, Yijun Yang, Huihui Xu, Weiming Wang, Jinni Zhou, and Lei Zhu.
\newblock Rainmamba: Enhanced locality learning with state space models for
  video deraining.
\newblock In {\em Proceedings of the 32nd ACM International Conference on
  Multimedia}, pages 7881--7890, 2024.

\bibitem{charbonnier1997deterministic}
Pierre Charbonnier, Laure Blanc-F{\'e}raud, Gilles Aubert, and Michel Barlaud.
\newblock Deterministic edge-preserving regularization in computed imaging.
\newblock {\em IEEE Transactions on image processing}, 6(2):298--311, 1997.

\bibitem{zhang2018unreasonable}
Richard Zhang, Phillip Isola, Alexei~A Efros, Eli Shechtman, and Oliver Wang.
\newblock The unreasonable effectiveness of deep features as a perceptual
  metric.
\newblock In {\em Proceedings of the IEEE conference on computer vision and
  pattern recognition}, pages 586--595, 2018.

\bibitem{paszke2017automatic}
Adam Paszke, Sam Gross, Soumith Chintala, Gregory Chanan, Edward Yang, Zachary
  DeVito, Zeming Lin, Alban Desmaison, Luca Antiga, and Adam Lerer.
\newblock Automatic differentiation in pytorch.
\newblock 2017.

\bibitem{kingma2014adam}
Diederik~P Kingma.
\newblock Adam: A method for stochastic optimization.
\newblock {\em arXiv preprint arXiv:1412.6980}, 2014.

\bibitem{wen2020ua}
Longyin Wen, Dawei Du, Zhaowei Cai, Zhen Lei, Ming-Ching Chang, Honggang Qi,
  Jongwoo Lim, Ming-Hsuan Yang, and Siwei Lyu.
\newblock Ua-detrac: A new benchmark and protocol for multi-object detection
  and tracking.
\newblock {\em Computer Vision and Image Understanding}, 193:102907, 2020.

\bibitem{hu2021v2e}
Yuhuang Hu, Shih-Chii Liu, and Tobi Delbruck.
\newblock v2e: From video frames to realistic dvs events.
\newblock In {\em Proceedings of the IEEE/CVF conference on computer vision and
  pattern recognition}, pages 1312--1321, 2021.

\bibitem{wang2006modern}
Zhou Wang and Alan~Conrad Bovik.
\newblock Modern image quality assessment.
\newblock 2006.

\bibitem{wang2004image}
Zhou Wang, Alan~C Bovik, Hamid~R Sheikh, and Eero~P Simoncelli.
\newblock Image quality assessment: from error visibility to structural
  similarity.
\newblock {\em IEEE transactions on image processing}, 13(4):600--612, 2004.

\bibitem{mittal2012making}
Anish Mittal, Rajiv Soundararajan, and Alan~C Bovik.
\newblock Making a “completely blind” image quality analyzer.
\newblock {\em IEEE Signal processing letters}, 20(3):209--212, 2012.

\end{thebibliography}
}
\end{document}